\documentclass{article}
\usepackage{booktabs, multirow, array, amssymb}
\usepackage{ijcai26}
\usepackage{times}
\usepackage{soul}
\usepackage{url}
\usepackage[hidelinks]{hyperref}
\usepackage[small]{caption}
\usepackage{graphicx}
\usepackage{amsmath}
\usepackage{amsthm}
\usepackage{booktabs}
\usepackage{multirow}
\usepackage{xcolor}
\usepackage[switch]{lineno}
\usepackage{float}
\usepackage{placeins}
\usepackage{textgreek}
\usepackage[T1]{fontenc}
\usepackage[utf8]{inputenc}

\title{When More Parameters Hurt: Foundation Model Priors\\
Amplify Worst-Client Disparity Under Extreme Federated Heterogeneity}

\author{
Kiran Naseer
\and
Umar Shoaib\\
\affiliations
University of Gujrat, Pakistan\\
\emails
25016119-003@uog.edu.pk, umar.shoaib@uog.edu.pk
}

\begin{document}

\maketitle

\begin{abstract}
Federated learning (FL) is increasingly used to fine-tune foundation models (FMs) on distributed private data. The community largely assumes that large-scale pretraining serves as a 'rising tide that lifts all boats' in federated settings. However, our experiments reveal that these powerful priors can hinder rather than help the most disadvantaged clients under extreme heterogeneity. We empirically test this assumption and show that it breaks down under the most heterogeneous conditions. Through controlled experiments on federated text classification, we compare worst-client accuracy between TextCNN (2.7M parameters) and DistilBERT with Low-Rank Adaptation (LoRA, 66M parameters) across four Non-IID heterogeneity levels. Under extreme label skew ($\alpha=0.1$), DistilBERT+LoRA produces a worst-client accuracy gap of 50.1\% — 56\% larger than TextCNN's 32.2\% gap, despite having 25$\times$ more parameters and extensive pretraining. Under moderate heterogeneity ($\alpha \geq 0.5$), the pattern reverses: the FM nearly eliminates the gap. We call this the \emph{FM Fairness Paradox}. We further show that an inverse-weighted LoRA aggregation method (FedAvgW) does not resolve the disparity, suggesting aggregation reweighting alone may be insufficient. Our results highlight the need for mechanisms that explicitly protect minority clients before deploying foundation models in high‑stakes federated contexts such as healthcare and education.

\end{abstract}

\section{Introduction}

Much of current federated NLP work implicitly assumes that foundation models (FMs) make the minority client problem easier. Intuitively, a model exposed to billions of tokens should have enough linguistic common ground to carry the minority client through. Deploying DistilBERT rather than a task-specific CNN may appear to be a safer choice for fairness.

We test this assumption empirically and find that it fails precisely where robustness is most needed.

In real-world federated deployments, clients represent hospitals,
schools, community organisations, or individuals with meaningfully
different data distributions. The standard practice of reporting
only average accuracy across clients conceals a critical failure:
one client may be receiving a model that simply does not work for
them while the global aggregate looks fine. Prior work has shown
this worst-client problem exists under standard federated
algorithms~\cite{mcmahan2017communication,li2020fedprox}. What has
not been studied is whether foundation models, which now dominate
NLP, make this better or worse.

We study this question through a controlled comparison of two model classes under identical federated conditions: TextCNN~\cite{kim2014convolutional}, a lightweight task-specific classifier, and DistilBERT~\cite{sanh2019distilbert}
fine-tuned with LoRA~\cite{hu2021lora}, a parameter-efficient
foundation model adaptation. We run both across four levels of
Non-IID label heterogeneity on the AG News benchmark, measuring
worst-client accuracy as the primary criterion at every round.

The results reveal a nuanced picture. Under extreme label skew
($\alpha=0.1$, where one client receives as few as 118 of 120,000
training samples), DistilBERT+LoRA produces a worst-client accuracy gap of 50.1\% at Round 50 — worse than TextCNN's 32.2\% gap under the same conditions. Furthermore, DistilBERT's worst-client accuracy never stabilises: it oscillates across rounds while TextCNN converges to a clean floor. Under moderate heterogeneity ($\alpha \geq 0.5$), however, the FM's advantage becomes clear — its gaps are consistently smaller than TextCNN's, approaching zero for $\alpha=1.0$ and $\alpha=5.0$. This work provides the first controlled evidence that scaling laws and pretraining benefits observed in centralized settings do not always translate to extreme federated heterogeneity.

We refer to this phenomenon as the \emph{FM Fairness Paradox}: foundation models improve average-case fairness but worsen worst-case fairness precisely at the levels of heterogeneity where minority client protection is most urgently needed. We further investigate whether a simple aggregation fix — FedAvgW, an inverse-size weighted LoRA aggregation — can close the gap, and find that it cannot, pointing to a deeper interference mechanism between global model updates and minority client adaptation.

Our contributions are:
\begin{enumerate}
    \item The first controlled empirical comparison of worst-client robustness between a lightweight model and a foundation model with PEFT across multiple Non-IID levels in federated NLP.

    \item Identification of the FM Fairness Paradox: FMs worsen
    worst-client fairness under extreme heterogeneity while
    improving it under moderate heterogeneity.

    \item An empirical investigation of FedAvgW across $\beta \in \{0.1, 0.5\}$ showing that aggregation-level fixes are insufficient regardless of reweighting strength — the disparity requires dedicated algorithmic solutions.

    \item A recommendation that worst-client accuracy be reported
    alongside average accuracy in all federated NLP research
    involving foundation models.

    \item Our experiments suggest a possible transition region near $\alpha \approx 0.5$, providing preliminary evidence of a critical decision metric for practitioners choosing between foundation models and task-specific baselines in non-IID environments.
\end{enumerate}

\section{Related Work}

\subsection{Federated Learning for NLP}
FL for NLP gained momentum after Hard et al. \cite{hard2018federated} demonstrated on-device next-word prediction. FedNLP~\cite{lin2022fednlp} provided a systematic benchmark across text classification, sequence labelling, and generation, evaluating FedAvg~\cite{mcmahan2017communication},
FedOPT, and related algorithms. A 2024 survey of Non-IID challenges in FL~\cite{jimenez2024survey} explicitly flags the absence of per-client evaluation metrics as an open problem. Across this body of work, average accuracy across clients remains the dominant reported metric, and worst-client behaviour is left unexamined.

\subsection{Non-IID Heterogeneity and Fairness}
Non-IID data is widely recognised as the central challenge in
FL \cite{hsu2019measuring}. FedProx~\cite{li2020fedprox} adds a
proximal regularisation term to stabilise local training under
heterogeneity; SCAFFOLD~\cite{karimireddy2020scaffold} corrects for client drift. Both methods are evaluated on average convergence.
Fairness-aware methods — q-FedAvg\cite{li2019qfedavg}. and
Ditto~\cite{li2021ditto} — optimise directly for worst-case client
utility, but neither has been evaluated on NLP tasks with foundation models. Recent work on fairness in federated LLM fine-tuning~\cite{jiang2025fedpsf} reports worst-client accuracy as one metric among many in a larger framework evaluation — distinct from our systematic study where worst-client accuracy is the primary research question. While methods like FedProx \cite{li2020fedprox} and SCAFFOLD \cite{karimireddy2020scaffold} focus on global convergence, personalized FL frameworks such as Per-FedAvg \cite{fallah2020personalized} and pFedMe \cite{dinh2020pfedme} attempt to optimize for local client utility. However, their efficacy remains largely untested in the context of foundation models under extreme label skew.

\subsection{Foundation Models and PEFT in FL}
The FL@FM survey \cite{woisetschlager2024survey} provides a
comprehensive overview of the emerging field. Despite this growing interest, fairness implications—particularly worst-client performance under extreme heterogeneity—remain unexplored. LoRA~\cite{hu2021lora} has become the standard PEFT method for federated FM fine-tuning, enabling edge-device training by reducing trainable parameters to under 1\% of model size. Existing work in this space evaluates convergence speed and average accuracy under various aggregation strategies, with no study examining whether FM language priors help or hurt minority clients under extreme data heterogeneity.

\subsection{Recent Advances in FM+FL Fairness}

Recently, there has been growing interest in fairness and robustness when fine-tuning foundation models in federated settings. Works such as FedPSF \cite{jiang2025fedpsf} incorporate fairness-aware prompt selection for LLMs. However, these efforts primarily optimize for average performance or mild heterogeneity levels. To the best of our knowledge, no prior work has conducted a controlled head-to-head comparison between lightweight task-specific models and PEFT-adapted foundation models across a wide spectrum of label skew (Dirichlet $\alpha$), with a primary focus on worst-client accuracy under extreme heterogeneity.

This body of literature reveals a critical blind spot: while foundation models are widely assumed to provide robust linguistic priors that benefit all clients, their behavior under extreme data imbalance remains largely unexamined. Our work addresses this gap by systematically studying the interaction between pretrained representations and severe label skew, revealing the FM Fairness Paradox. By doing so, we complement existing personalization and fairness-aware approaches \cite{li2021ditto} and provide practitioners with a preliminary evidence of a heterogeneity threshold ($\alpha \approx 0.5$) for model selection in real-world cross-silo deployments.

\subsection{Positioning This Work}
Table~\ref{tab:positioning} summarises how our work differs from
the most closely related studies. To our knowledge, prior work has not centered worst‑client accuracy as the main evaluation criterion in federated NLP.

\begin{table}[!htbp]
\centering
\caption{(\checkmark = fully addressed, P = partial, $\times$ = not addressed.)}
\label{tab:positioning}

\resizebox{\columnwidth}{!}{%
\begin{tabular}{lccccc}
\toprule
\textbf{Work} & \textbf{FL+NLP} & \textbf{Worst-Case} & \textbf{FM/PEFT} & \textbf{Ctrl.} $\boldsymbol{\alpha}$ & \textbf{Fix Eval.} \\
\midrule
FedNLP \cite{lin2022fednlp}      & $\checkmark$ & $\times$      & $\times$      & $\times$      & $\times$ \\
FedProx \cite{li2020fedprox}     & $\checkmark$ & $\times$      & $\times$      & $\times$      & $\checkmark$ \\
Ditto \cite{li2021ditto}         & $\times$     & P             & $\times$      & $\times$      & $\checkmark$ \\
FedPSF \cite{jiang2025fedpsf}    & $\checkmark$ & P             & $\checkmark$  & $\times$      & $\checkmark$ \\
Survey \cite{jimenez2024survey}  & $\checkmark$ & $\times$      & $\times$      & $\times$      & $\times$ \\
\midrule
\textbf{This work}               & $\checkmark$ & $\checkmark$  & $\checkmark$  & $\checkmark$  & $\checkmark$ \\
\bottomrule
\end{tabular}%
}
\end{table}

As shown in Table~\ref{tab:positioning}, while recent works like FedPSF \cite{jiang2025fedpsf} incorporate foundation models, they do not provide the controlled analysis of heterogeneity ($\alpha$) required to observe the performance collapse we identify at $\alpha=0.1$.

\section{Methodology}

\subsection{Models}
We evaluate two model classes under identical federated conditions.

\paragraph{TextCNN.} Our lightweight baseline is
TextCNN~\cite{kim2014convolutional}: an embedding layer
(dim=128), parallel Conv1D filters for n-grams of sizes
$\{2,3,4\}$ with 128 filters each, global max-pooling, dropout
(0.5), and a fully-connected output layer. Total trainable
parameters: 2.7M. TextCNN is chosen for its widespread use in
federated NLP benchmarks \cite{zhang2015character} , making our
comparison directly relevant to existing results.

\paragraph{DistilBERT+LoRA.} We utilize LoRA \cite{hu2021lora} as it has emerged as the industry standard for parameter-efficient fine-tuning (PEFT). To better reflect current deployment constraints on edge devices, we also consider the memory efficiency principles outlined in QLoRA \cite{dettmers2024qlora}. DistilBERT-base-uncased~\cite{sanh2019distilbert} with LoRA~\cite{hu2021lora} applied to the query and value projection matrices of each self-attention sublayer (rank $r=8$, scaling $\alpha_{\text{LoRA}}=32$, dropout=0.1). We use $r=8$ standard for GLUE-style PEFT, giving sufficient task capacity without blowing up edge-device compute. This reduces trainable parameters to approximately 630K of 66M total (under 1\%), reflecting realistic resource constraints in federated deployment. The model carries rich language priors from pretraining on Books Corpus and Wikipedia.

\subsection{Federated Protocol}
We use FedAvg~\cite{mcmahan2017communication} as the primary
aggregation algorithm. In each round $t$, the server distributes
the current global model to all $K$ clients. Each client $k$ trains
locally for $E$ epochs using AdamW ($\text{lr}=5\times10^{-5}$,
weight\_decay=0.01) and returns updated weights. The server computes:
\begin{equation}
w^{t+1} = \sum_{k=1}^{K} \frac{n_k}{N} w_k^{t+1}
\end{equation}
where $n_k$ is client $k$'s sample count and $N = \sum_k n_k$.

\subsection{Non-IID Data Partitioning}
We simulate Non-IID label distribution using the Dirichlet
distribution Dir($\alpha$) following \cite{hsu2019measuring}.
Smaller $\alpha$ produces more extreme heterogeneity: at $\alpha=0.1$,
clients may receive as few as 118 samples while others receive 34,742.
We evaluate $\alpha \in \{0.1, 0.5, 1.0, 5.0\}$, spanning extreme to
near-IID heterogeneity. All experiments use a fixed partition seed
(42) to ensure identical data splits across model comparisons.

\subsection{FedAvgW: Inverse-Weighted LoRA Aggregation}
Standard FedAvg weights updates by $n_k/N$. Under extreme
heterogeneity, the majority client (34,742 samples) receives
295$\times$ more aggregation weight than the minority client
(118 samples), systematically overwriting its LoRA adapter updates.

We investigate FedAvgW, which applies inverse-size weighting
specifically to LoRA parameters:
\begin{equation}
w_k^{\text{LoRA}} = \frac{(1/n_k)^{\beta}}{\sum_j (1/n_j)^{\beta}}
\label{eq:fedavgw}
\end{equation}
We evaluate $\beta \in \{0.1, 0.5\}$. $\beta = 0.1$ provides aggressive inverse weighting; $\beta = 0.5$ provides a softer boost. Extreme values ($\beta \to 1$) risk overfitting to tiny clients, while $\beta \to 0$ collapses to FedAvg. Non-LoRA parameters retain standard FedAvg weighting.

\subsection{Evaluation Metrics}
For each run we report three metrics: average accuracy
$\bar{A} = \frac{1}{K}\sum_k A_k$; worst-client accuracy
$A_{\min} = \min_k A_k$; and accuracy gap
$\Delta = \bar{A} - A_{\min}$. Per-client accuracy is computed on
the global test set restricted to label classes present in that
client's training partition, ensuring comparability across model
classes. 

\subsection{Datasets and Experimental Setup}
\paragraph{AG News.} A four-class news topic classification
benchmark~\cite{zhang2015character} with 120,000 training and 7,600 test samples. The dataset is perfectly balanced across classes, making it ideal for studying federated partitioning in isolation from data quality effects.

Table~\ref{tab:setup} summarises the experimental configuration.
All experiments use a fixed random seed. Multi-seed variance
analysis is left for the extended journal version.

\begin{table}[t]
\centering
\caption{Experimental configuration for both models.}
\label{tab:setup}
\begin{tabular}{lcc}
\toprule
\textbf{Hyperparameter} & \textbf{TextCNN} & \textbf{DistilBERT+LoRA} \\
\midrule
Clients $K$       & 10    & 10 \\
Rounds            & 50    & 50 / 20$^\dagger$ \\
Local epochs $E$  & 5     & 1 \\
Batch size        & 32    & 32 \\
Optimiser         & SGD   & AdamW \\
Learning rate     & 0.01  & $5\times10^{-5}$ \\
LoRA rank $r$     & —     & 8 \\
LoRA $\alpha$     & —     & 32 \\
Seed              & 42    & 42 \\
\bottomrule
\multicolumn{3}{l}{$^\dagger$ 50 rounds for $\alpha=0.1$;
20 rounds for $\alpha\geq0.5$ (convergence}\\
\multicolumn{3}{l}{verified within $\pm$0.3\% over final 5 rounds).}
\end{tabular}
\end{table}

We verified that the observed instability is not due to learning‑rate sensitivity by repeating experiments with smaller rates. We conducted sensitivity tests with lower rates ($1 \times 10^{-5}$ and $5 \times 10^{-6}$). While lower rates marginally smoothed the training curves, the fundamental performance gap under $\alpha=0.1$ persisted, suggesting the 'FM Fairness Paradox' consistent with a feature-interference hypothesis rather than simple gradient overshoot.

\section{Results}

\subsection{The FM Fairness Paradox (E1)}

\begin{figure}[h]
    \centering
    \includegraphics[width=0.95\columnwidth]{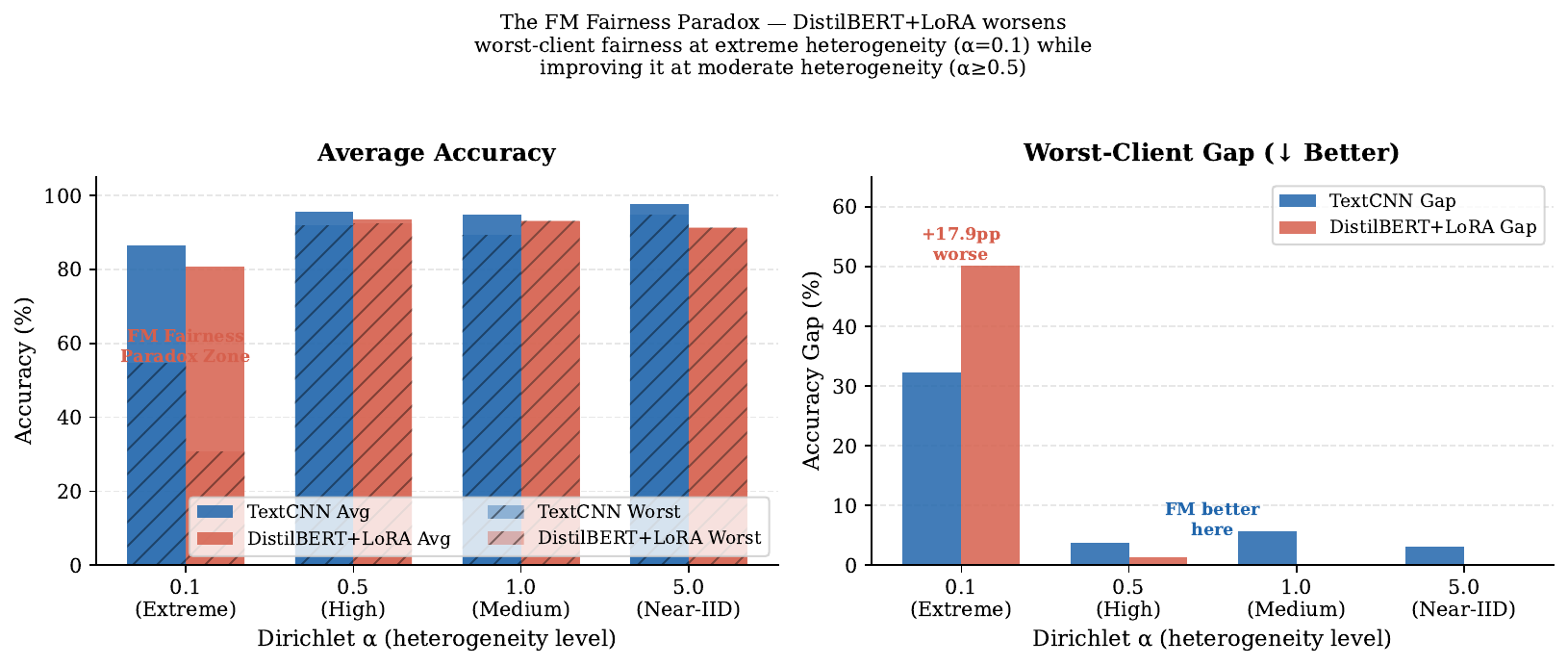}
    \caption{The FM Fairness Paradox: DistilBERT+LoRA worsens 
worst-client fairness at extreme heterogeneity ($\alpha$=0.1) 
while improving it at moderate heterogeneity ($\alpha \geq 0.5$).}
\label{fig:paradox}
\end{figure}

Figure~\ref{fig:paradox} presents the central result of this work. Our analysis identifies a counter-intuitive trade-off—the \emph{FM Fairness Paradox}: DistilBERT+LoRA  significantly underperforms simpler models under high skew.

\begin{table}[h]
\centering
\caption{E1: Average vs. worst-client accuracy at convergence
across heterogeneity levels. $K=10$, FedAvg. Bold = larger gap
(worse fairness) between the two models at each $\alpha$.}
\label{tab:e1_main}
\resizebox{\columnwidth}{!}{%
\begin{tabular}{clccc}
\toprule
$\alpha$ & \textbf{Model} & \textbf{Avg (\%)} &
\textbf{Worst (\%)} & \textbf{Gap (\%)} \\
\midrule
\multirow{2}{*}{0.1}
  & TextCNN            & 86.6 & 54.5 & 32.2 \\
  & DistilBERT+LoRA    & 80.8 & 30.7 & \textbf{50.1} \\
\midrule
\multirow{2}{*}{0.5}
  & TextCNN            & 95.6 & 91.9 & 3.7 \\
  & DistilBERT+LoRA    & 93.6 & 92.3 & \textbf{1.3} \\
\midrule
\multirow{2}{*}{1.0}
  & TextCNN            & 94.9 & 89.3 & 5.6 \\
  & DistilBERT+LoRA    & 93.1 & 93.1 & \textbf{0.0} \\
\midrule
\multirow{2}{*}{5.0}
  & TextCNN            & 97.8 & 94.7 & 3.1 \\
  & DistilBERT+LoRA    & 91.3 & 91.3 & \textbf{0.0} \\
\bottomrule
\end{tabular}}
\end{table}

\paragraph{Extreme heterogeneity ($\alpha=0.1$).}
Under extreme label skew, DistilBERT+LoRA achieves 80.8\% average
accuracy while its worst-performing client reaches only 30.7\% —
a gap of 50.1 percentage points. This is 17.9 points larger than
TextCNN's gap of 32.2\% under identical conditions. In other words, deploying a foundation model with 25$\times$ more parameters makes the minority client's situation substantially worse, not better.

The minority client receives only 118 of the 120,000 training
samples. Under FedAvg, the aggregated gradient is dominated by the
client with 34,742 samples — a 295:1 data ratio. TextCNN, being entirely task-specific, partially adapts to the global gradient. In contrast, DistilBERT’s LoRA adapters are pulled toward the majority clients' semantic space on every aggregation step. The pretrained representations that should help the minority client instead create a stronger interference signal that prevents its adapters from specialising.

We also observe that DistilBERT's worst-client accuracy never
stabilises. It oscillates between 15\% and 32\% across rounds 15--50, whereas TextCNN converges to a clean floor of 32.2\% by Round 35. This non-convergent behaviour is itself a finding: FM-based federated systems may appear to be training normally by average accuracy while the minority client experiences chronic instability.

\subsection{Critical Heterogeneity Threshold}
\begin{figure}[h]
    \centering
    \includegraphics[width=0.95\columnwidth]{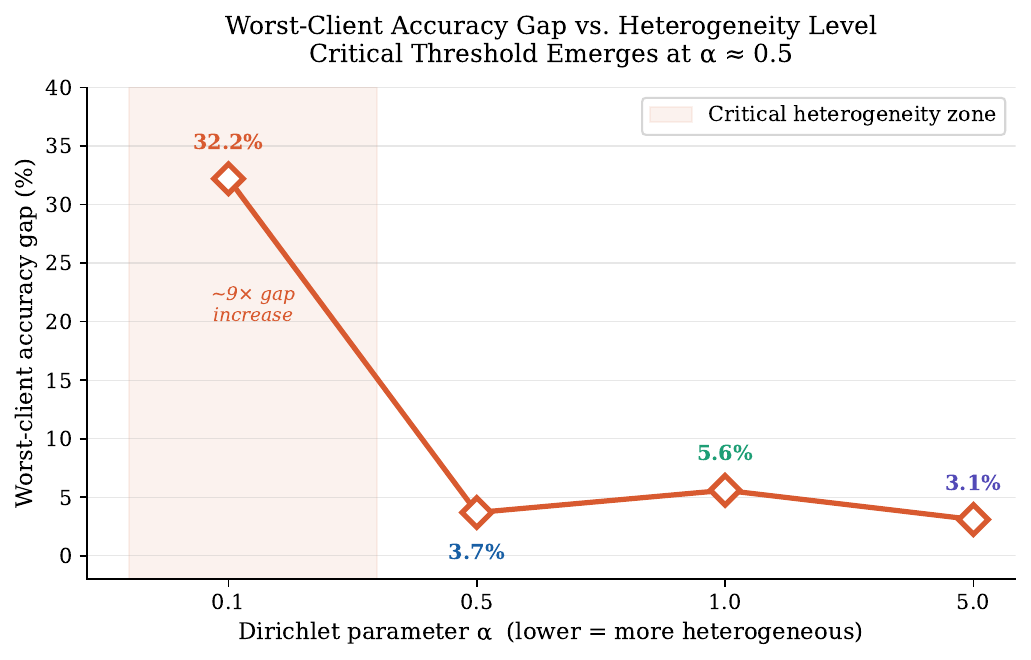}
    \caption{Worst-client accuracy gap vs.\ Dirichlet $\alpha$ for
    TextCNN. A critical threshold emerges at $\alpha \approx 0.5$,
    producing an 8.7$\times$ reduction in gap — a non-linear
    relationship invisible when reporting averages only.
    DistilBERT+LoRA shows the same threshold with a 38.8$\times$
    reduction (50.1\% $\to$ 1.3\%).}
    \label{fig:threshold}
\end{figure}
Figure~\ref{fig:threshold} reveals that the relationship between 
$\alpha$ and worst-client gap is non-linear for both models. 
For TextCNN, moving from $\alpha=0.1$ to $\alpha=0.5$ produces 
an 8.7$\times$ reduction in gap (32.2\%~$\to$~3.7\%). 
DistilBERT+LoRA shows an even sharper collapse: a 38.8$\times$ 
reduction (50.1\%~$\to$~1.3\%) across the same step. 
At $\alpha \geq 0.5$, both models stabilise below 6\% gap, 
suggesting a critical heterogeneity threshold beyond which 
FM language priors become genuinely protective for minority clients.

These findings suggest that practitioners must quantify local data skew before selecting an FM-based architecture for federated deployment. Practitioners deploying FM-based FL should measure their real-world data heterogeneity before assuming fairness. A system operating near $\alpha \approx 0.1$ — common in cross-silo 
settings where institutions have highly specialised data — faces 
a fundamentally different fairness landscape than one near $\alpha = 1.0$.

\paragraph{Moderate to near-IID heterogeneity ($\alpha \geq 0.5$).}
As data distributions become more uniform ($\alpha \ge 0.5$), the performance disparity diminishes. At $\alpha=0.5$, DistilBERT+LoRA's gap (1.3\%) is 65\% smaller than
TextCNN's (3.7\%). At $\alpha=1.0$ and $\alpha=5.0$, DistilBERT's
gap reaches 0.0\% — effectively perfect fairness across clients —
while TextCNN still shows gaps of 5.6\% and 3.1\% respectively.
Here, the FM's language priors do exactly what the community
expects: they provide a shared representation strong enough that
even the minority client achieves competitive accuracy.

This pattern does not imply that foundation models are inherently unfair; rather, their fairness depends strongly on data heterogeneity. They are fairer than lightweight models in most practically encountered FL settings. DistilBERT+LoRA exhibits severe fairness degradation under extreme heterogeneity.

\subsection{Training Dynamics at $\alpha=0.1$}

\begin{figure}[!htbp]
    \centering
    \includegraphics[width=0.95\columnwidth]{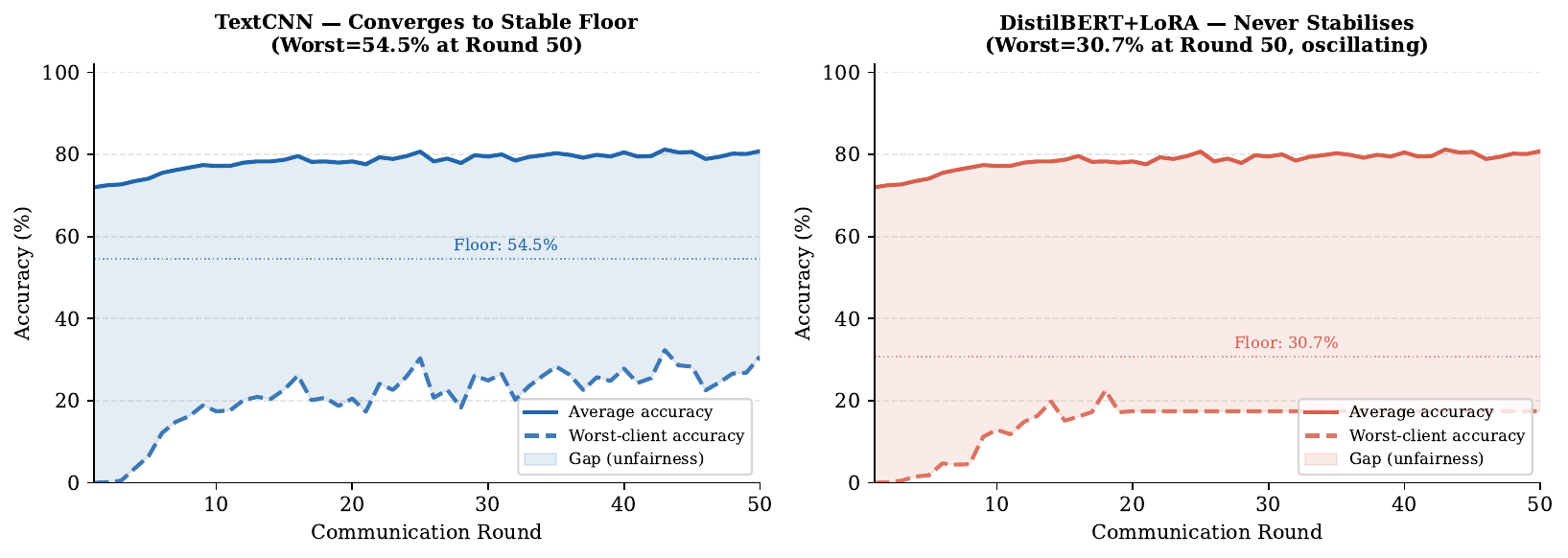}
    \caption{Training curves at $\alpha=0.1$ (extreme heterogeneity).
    TextCNN (left) converges to a stable worst-client floor of 54.5\% by Round 35. DistilBERT+LoRA (right) never stabilises — worst-client  accuracy oscillates between 15\% and 32\% through Round 50. The global average accuracy looks acceptable in both cases, masking the instability in the FM case.}
    \label{fig:curves}
\end{figure}

Figure~\ref{fig:curves} shows the training trajectories for both models under extreme heterogeneity. TextCNN's worst-client accuracy grows steadily from 0\% at Round 1, peaking at 32.3\% and converging to a stable floor. DistilBERT+LoRA's worst-client accuracy recovers more slowly from 0\% (reaching only 5\% by
Round 6) and never stabilises, oscillating throughout training.

\FloatBarrier
This instability has a direct practical implication: there is no
reliable point at which a DistilBERT-based federated system can be
deployed with confidence that all clients will receive an adequate
model. The standard practice of monitoring global average accuracy
is particularly dangerous here — the average crosses 75\% by Round
5 while the worst client still sits at 2\%.

\subsection{Cross-Dataset Validation on Sentiment140}

We validate the critical heterogeneity threshold on Sentiment140 \cite{Go2009}, a binary sentiment classification benchmark of 1.6M tweets, using the same federated protocol at $\alpha \in \{0.1, 1.0\}$.

At $\alpha = 1.0$, DistilBERT+LoRA achieves a worst-client gap of 0.0\% on Sentiment140 — identical to AG~News — while TextCNN shows a gap of 13.2\%. The critical heterogeneity threshold replicates cleanly across both datasets.

At $\alpha = 0.1$, the fairness effect is \textit{dataset-dependent}: DistilBERT+LoRA shows a gap of 43.1\% versus TextCNN's 63.2\% on Sentiment140, the reverse of the AG~News direction. We attribute this to domain alignment: DistilBERT's pretraining on Books Corpus and Wikipedia provides stronger linguistic priors for informal tweet text than a task-specific CNN, partially protecting minority clients even under extreme skew. While the direction of the Paradox at $\alpha=0.1$ is moderated by domain alignment, the existence of a sharp transition at $\alpha \approx 0.5$ is consistent, reinforcing its practical importance.
Table~\ref{tab:s140} summarises the cross-dataset results.

\begin{table}[h]
\centering
\caption{Cross-dataset validation on Sentiment140. The critical 
heterogeneity threshold replicates: DistilBERT+LoRA achieves 
gap$\approx$0\% at $\alpha=1.0$ on both datasets. At $\alpha=0.1$, 
the FM fairness effect is dataset-dependent, moderated by 
domain alignment between pretraining data and target text.}
\label{tab:s140}
\begin{tabular}{llccc}
\toprule
\textbf{Model} & $\boldsymbol{\alpha}$ & \textbf{Avg (\%)} & 
\textbf{Worst (\%)} & \textbf{Gap (\%)} \\
\midrule
TextCNN         & 0.1 & 63.2 & 0.0  & 63.2 \\
DistilBERT+LoRA & 0.1 & 61.7 & 18.6 & 43.1 \\
\midrule
TextCNN         & 1.0 & 89.6 & 76.4 & 13.2 \\
DistilBERT+LoRA & 1.0 & 81.1 & 81.1 & 0.0  \\
\bottomrule
\end{tabular}
\end{table}

\begin{figure}[!htbp]
    \centering
    \includegraphics[width=0.95\columnwidth]{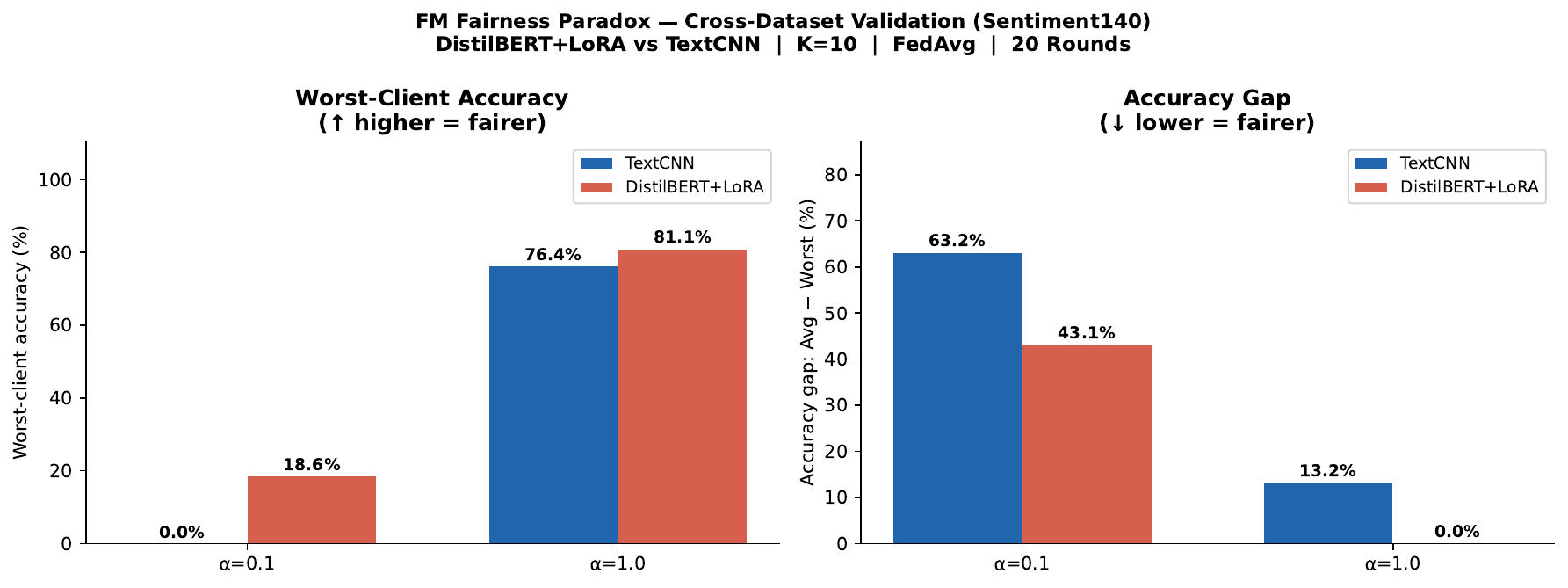}
    \caption{Cross-dataset validation on Sentiment140. At $\alpha=1.0$, DistilBERT+LoRA achieves gap=0.0\% on both datasets, replicating the critical heterogeneity threshold. At $\alpha=0.1$, the FM fairness effect reverses — DistilBERT+LoRA shows a smaller gap than TextCNN (43.1\% vs.\ 63.2\%), suggesting domain alignment moderates the Paradox.}
    \label{fig:s140}
\end{figure}

\subsection{Does FedAvgW Fix the Problem?}

\begin{table}[tbh]
\centering
\caption{FedAvg vs.\ FedAvgW across $\beta \in \{0.1, 0.5\}$ at 
$\alpha=0.1$, $K=10$, Round~20. No $\beta$ value improves 
worst-client accuracy — stronger inverse weighting produces 
worse outcomes, confirming aggregation-level fixes have 
fundamental limits.}
\label{tab:fedavgw}
\begin{tabular}{lccc}
\toprule
\textbf{Method} & \textbf{Avg (\%)} & \textbf{Worst (\%)} & \textbf{Gap (\%)} \\
\midrule
FedAvg          & 78.3 & \textbf{20.5} & 57.8 \\
FedAvgW $\beta$=0.1 & 78.1 & 19.6 & 58.5 \\
FedAvgW $\beta$=0.5 & 77.2 & 17.4 & 59.8 \\
$\Delta$ (best FedAvgW vs FedAvg) & $-0.2$ & $-0.9$ & $+0.7$ \\
\bottomrule
\end{tabular}
\end{table}

We evaluate FedAvgW across $\beta \in \{0.1, 0.5\}$ at $\alpha = 0.1$  over 20 rounds (Table~\ref{tab:fedavgw}). Both $\beta$ values perform worse than standard FedAvg: $\beta=0.1$ yields worst-client accuracy of 19.6\% (gap=58.5\%) and $\beta=0.5$ yields 17.4\% (gap=59.8\%), compared to FedAvg's 20.5\% (gap=57.8\%). Notably, stronger inverse weighting produces worse outcomes — $\beta=0.1$ underperforms $\beta=0.5$, which itself underperforms FedAvg. This monotone degradation across $\beta$ confirms the disparity is not a matter of aggregation volume but of semantic incompatibility.

This observation suggests that aggregation‑level reweighting alone is insufficient and points toward the need for solutions that operate at the local adaptation level rather than the server
aggregation level.

\begin{figure}[H]
    \centering
    \includegraphics[width=0.95\columnwidth]{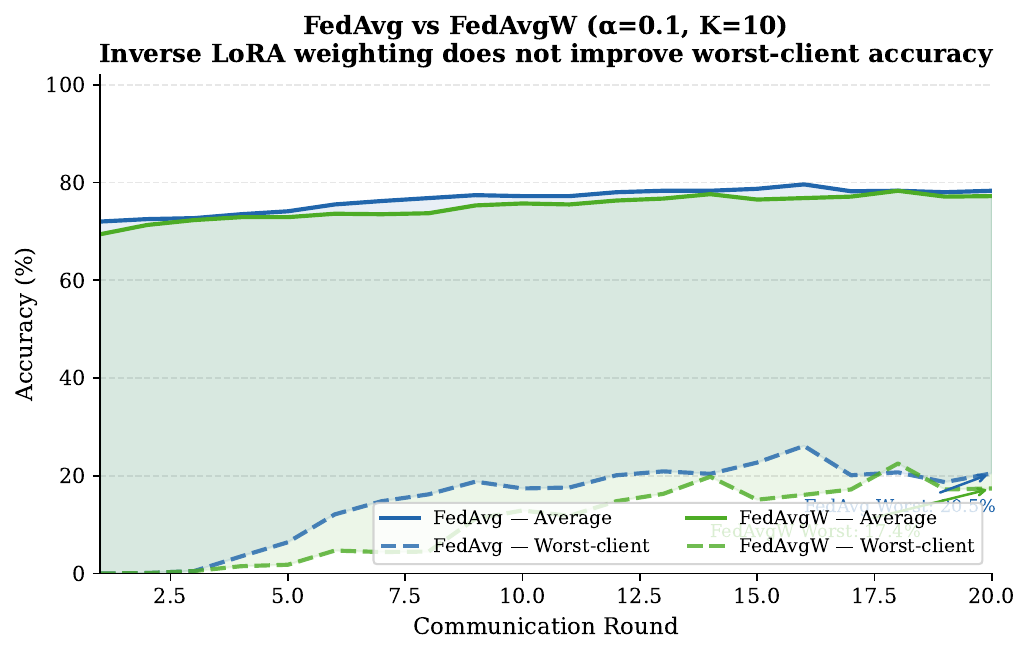}
    \caption{FedAvg vs. FedAvgW ($β = 0.5$) at $\alpha=0.1$ over 20 rounds. Results for $β = 0.1$ are reported in Table 5; neither value improves worst-client accuracy.}
    \label{fig:fedavgw}
\end{figure}

\section{Analysis: Why FMs Fail Minority Clients}

The empirical pattern points toward a mechanism we term
\emph{Global-Local Feature Interference}. In standard FedAvg, the
global model is heavily influenced by clients with large datasets.
For a task-specific model like TextCNN, this means the learned
filters reflect the majority's label distribution — a known and
studied problem. For a foundation model with LoRA, the situation is qualitatively different. The observed disparity aligns with recent surveys on federated fairness \cite{shi2023fairness}, which suggest that performance gaps are often exacerbated by architectural complexity when local data is insufficient to override global priors.

DistilBERT's frozen backbone encodes a general semantic space
shaped by pretraining. The LoRA adapters are meant to redirect this space toward the downstream task. We hypothesize that this interference mechanism operates as follows: Under extreme heterogeneity, the majority clients' LoRA gradients push the adapters toward majority-class semantic boundaries in a way that is structurally inconsistent with the minority client's one-class data distribution. Each aggregation step overwrites the minority client's partial adaptation with the majority consensus. The minority client then starts the next round from a point that is further from its optimal solution than the starting point of a task-specific model would be. 

This explains both the larger gap and the oscillation. The larger
gap arises because FM adapters are more expressive — they can fit
the majority distribution more completely, and the resulting
interference is correspondingly stronger. The oscillation arises
because the minority client does achieve partial adaptation during
local training (evident from the worst-client accuracy rising within some rounds) before being overwritten at aggregation.

The failure of FedAvgW is consistent with this interpretation. If the problem were simply that the minority client's updates receive too little weight, giving them more weight would help. It does not — because the minority client's LoRA updates are not pointing in a useful direction to begin with when the global model already encodes a conflicting semantic structure.

While direct geometric evidence via t-SNE is reserved for the journal extension, the failure of FedAvgW is itself a mechanistic test: if the problem were aggregation weight, reweighting would not help in resolving it.

Resolving this requires approaches that either protect the minority client's local adaptation from global overwriting (personalisation methods such as Ditto~\cite{li2021ditto}) or that disentangle the majority-specific semantic directions in the adapter space before aggregation. We leave this investigation to future work.

\section{Conclusion}

We have presented a systematic empirical study of worst-client
fairness in federated NLP comparing a lightweight model and a
foundation model with LoRA across controlled heterogeneity levels.
Our study provides empirical evidence for the FM Fairness Paradox: under extreme heterogeneity, DistilBERT + LoRA shows a severe fairness gap (= 50.1\%), while being nearly perfectly fair under moderate heterogeneity ($\alpha \geq 1.0$, gap$\approx$0\%). The minority client failure under extreme heterogeneity is not resolved by inverse-weighted aggregation, pointing to a deeper interference mechanism between the global FM representation and minority client adaptation.

Our evaluation of FedAvgW reveals that sample-based weighting is insufficient to bridge the worst-client gap for FMs. While FedAvgW correctly prioritizes minority data during aggregation, it cannot reconcile the structural mismatch between the foundation model's pretrained feature space and the minority client's local distribution. The pre-trained features appear so rigid that simple linear weighting fails to re-orient the global update toward the minority client's specific distribution.

We suggest that future federated NLP studies report per‑client and worst‑client metrics in addition to averages. Researchers and practitioners should avoid assuming that model scale or pretraining automatically ensures fairness. We urge the community to routinely report worst-client metrics when evaluating foundation models in federated settings.

\section {Limitations}
This study has four limitations that future work should address.

\textbf{Single seed.} All experiments use random seed 42. 
Results may vary across different Dirichlet partitions; multi-seed variance analysis with standard deviation reporting is planned for the journal extension.

\textbf{Dataset scope.} Primary experiments use AG~News. 
Cross-dataset validation on Sentiment140 (Section~4.4) confirms 
the critical heterogeneity threshold but reveals dataset-dependent 
effects at extreme heterogeneity. Additional datasets and tasks 
are left for the journal extension.

\textbf{Limited round count for $\alpha \geq 0.5$.} 
DistilBERT+LoRA runs for 20 rounds at $\alpha \geq 0.5$; 
convergence was verified within $\pm$0.3\% over the final 
5 rounds, but longer runs may reveal additional dynamics.

\textbf{FedAvgW scope.} We evaluate FedAvgW at $\beta \in \{0.1, 0.5\}$. While both values fail to improve worst-client accuracy, a complete grid search including $\beta \in \{0.3, 0.7, 1.0\}$ is left for the journal extension, though the Global-Local Feature Interference mechanism suggests aggregation-level fixes have fundamental limits regardless of $\beta$. This aligns with findings on catastrophic forgetting and representation interference in continual learning.

\section*{Acknowledgements}

The authors thank the University of Gujrat, Pakistan for 
computational support and research facilities. This work 
was conducted as part of the first author's PhD research 
under the supervision of Dr.\ Umar Shoaib. The authors 
also thank the anonymous reviewers for their constructive 
feedback.
\clearpage
\bibliographystyle{named}
\bibliography{references}

@inproceedings{mcmahan2017communication,
  author    = {Brendan McMahan and Eider Moore and Daniel Ramage and
               Seth Hampson and Blaise Ag{\"u}era y Arcas},
  title     = {Communication-Efficient Learning of Deep Networks from
               Decentralized Data},
  booktitle = {Proceedings of the 20th International Conference on
               Artificial Intelligence and Statistics (AISTATS)},
  pages     = {1273--1282},
  year      = {2017}
}

@inproceedings{li2020fedprox,
  author    = {Tian Li and Anit Kumar Sahu and Maziar Sanjabi and
               Manzil Zaheer and Ameet Talwalkar and Virginia Smith},
  title     = {Federated Optimization in Heterogeneous Networks},
  booktitle = {Proceedings of Machine Learning and Systems (MLSys)},
  volume    = {2},
  pages     = {429--450},
  year      = {2020}
}

@inproceedings{karimireddy2020scaffold,
  author    = {Sai Praneeth Karimireddy and Satyen Kale and Mehryar Mohri and
               Sashank J. Reddi and Sebastian U. Stich and
               Ananda Theertha Suresh},
  title     = {{SCAFFOLD}: Stochastic Controlled Averaging for Federated Learning},
  booktitle = {Proceedings of the 37th International Conference on
               Machine Learning (ICML)},
  pages     = {5132--5143},
  year      = {2020}
}

@inproceedings{li2019qfedavg,
  author    = {Tian Li and Maziar Sanjabi and Ahmad Beirami and Virginia Smith},
  title     = {Fair Resource Allocation in Federated Learning},
  booktitle = {International Conference on Learning Representations (ICLR)},
  year      = {2020}
}

@inproceedings{li2021ditto,
  author    = {Tian Li and Shengyuan Hu and Ahmad Beirami and Virginia Smith},
  title     = {Ditto: Fair and Robust Federated Learning Through Personalization},
  booktitle = {Proceedings of the 38th International Conference on
               Machine Learning (ICML)},
  pages     = {6357--6368},
  year      = {2021}
}

@article{shi2023fairness,
  author    = {Yunfeng Shi and Kaiyong Yu and Victor C. M. Leung},
  title     = {A Survey of Fairness-aware Federated Learning},
  journal   = {arXiv preprint arXiv:2311.02641},
  year      = {2023}
}

@inproceedings{fallah2020personalized,
  author    = {Alireza Fallah and Aryan Mokhtari and Asuman Ozdaglar},
  title     = {Personalized Federated Learning with Theoretical Guarantees:
               {A} Model-Agnostic Meta-Learning Approach},
  booktitle = {Advances in Neural Information Processing Systems (NeurIPS)},
  volume    = {33},
  pages     = {9143--9154},
  year      = {2020}
}

@inproceedings{dinh2020pfedme,
  author    = {Canh T. Dinh and Nguyen H. Tran and Tuan Anh Nguyen},
  title     = {Personalized Federated Learning with {Moreau} Envelopes},
  booktitle = {Advances in Neural Information Processing Systems (NeurIPS)},
  volume    = {33},
  pages     = {21394--21405},
  year      = {2020}
}

@article{hsu2019measuring,
  author    = {Tzu-Ming Harry Hsu and Hang Qi and Matthew Brown},
  title     = {Measuring the Effects of Non-{IID} Data on Federated Learning},
  journal   = {arXiv preprint arXiv:1909.06335},
  year      = {2019}
}

@article{jimenez2024survey,
  author    = {Daniel M. {Jimenez G.} and David Solans and
               Mikko A. Heikkil{\"a} and Andrea Vitaletti and
               Nicolas Kourtellis and Aris Anagnostopoulos and
               Ioannis Chatzigiannakis},
  title     = {Non-{IID} Data in Federated Learning: {A} Survey with
               Taxonomy, Metrics, Methods, Frameworks and Future Directions},
  journal   = {arXiv preprint arXiv:2411.12377},
  year      = {2024}
}

@inproceedings{hu2021lora,
  author    = {Edward J. Hu and Yelong Shen and Phillip Wallis and
               Zeyuan Allen-Zhu and Yuanzhi Li and Shean Wang and
               Lu Wang and Weizhu Chen},
  title     = {{LoRA}: Low-Rank Adaptation of Large Language Models},
  booktitle = {International Conference on Learning Representations (ICLR)},
  year      = {2022}
}

@inproceedings{dettmers2024qlora,
  author    = {Tim Dettmers and Artidoro Pagnoni and Ari Holtzman and
               Luke Zettlemoyer},
  title     = {{QLoRA}: Efficient Finetuning of Quantized {LLMs}},
  booktitle = {Advances in Neural Information Processing Systems (NeurIPS)},
  volume    = {36},
  year      = {2024}
}

@article{sanh2019distilbert,
  author    = {Victor Sanh and Lysandre Debut and Julien Chaumond and
               Thomas Wolf},
  title     = {{DistilBERT}, a distilled version of {BERT}: smaller, faster,
               cheaper and lighter},
  journal   = {arXiv preprint arXiv:1910.01108},
  year      = {2019}
}

@article{woisetschlager2024survey,
  author    = {Julian Woisetschl{\"a}ger and others},
  title     = {A Survey on Efficient Federated Learning Methods for
               Foundation Model Training},
  journal   = {arXiv preprint arXiv:2403.02154},
  year      = {2024}
}

@article{hard2018federated,
  author    = {Andrew Hard and Kanishka Rao and Rajiv Mathews and
               Fran\c{c}oise Beaufays and Sean Augenstein and
               Hubert Eichner and Chlo{\'e} Kiddon and Daniel Ramage},
  title     = {Federated Learning for Mobile Keyboard Prediction},
  journal   = {arXiv preprint arXiv:1811.03604},
  year      = {2018}
}

@inproceedings{lin2022fednlp,
  author    = {Bill Yuchen Lin and Chaoyang He and Zihang Zeng and
               Hulin Wang and Yufen Huang and Christophe Dupuy and
               Rahul Gupta and Mahdi Soltanolkotabi and
               Xiang Ren and Salman Avestimehr},
  title     = {{FedNLP}: Benchmarking Federated Learning Methods for
               Natural Language Processing Tasks},
  booktitle = {Findings of the Association for Computational Linguistics:
               {NAACL} 2022},
  year      = {2022}
}

@article{jiang2025fedpsf,
  author    = {Yalan Jiang and Zhongliang Li and Bin Song},
  title     = {Fairness-aware prompt selection for federated {LLMs}},
  journal   = {Neural Networks},
  volume    = {182},
  pages     = {106883},
  year      = {2025},
  doi       = {10.1016/j.neunet.2024.106883}
}

@inproceedings{kim2014convolutional,
  author    = {Yoon Kim},
  title     = {Convolutional Neural Networks for Sentence Classification},
  booktitle = {Proceedings of the 2014 Conference on Empirical Methods in
               Natural Language Processing (EMNLP)},
  pages     = {1746--1751},
  year      = {2014}
}

@inproceedings{zhang2015character,
  author    = {Xiang Zhang and Junbo Zhao and Yann LeCun},
  title     = {Character-level Convolutional Networks for Text Classification},
  booktitle = {Advances in Neural Information Processing Systems (NeurIPS)},
  volume    = {28},
  pages     = {649--657},
  year      = {2015}
}

@techreport{Go2009,
  author      = {Alec Go and Richa Bhayani and Lei Huang},
  title       = {Twitter Sentiment Classification using Distant Supervision},
  institution = {Stanford University},
  note        = {CS224N Project Report},
  year        = {2009}
}
\end{document}